# Exploiting Synergy Between Ontologies and Recommender Systems

Stuart E. Middleton, Harith Alani, David C. De Roure
Intelligence, Agents and Multimedia Group
Department of Electronics and Computer Science
University of Southampton
Southampton, SO17 1BJ, UK

{sem99r,ha,dder}@ecs.soton.ac.uk

## ABSTRACT
Recommender systems learn about user preferences over time, automatically finding things of similar interest. This reduces the burden of creating explicit queries. Recommender systems do, however, suffer from cold-start problems where no initial information is available early on upon which to base recommendations.

Semantic knowledge structures, such as ontologies, can provide valuable domain knowledge and user information. However, acquiring such knowledge and keeping it up to date is not a trivial task and user interests are particularly difficult to acquire and maintain.

This paper investigates the synergy between a web-based research paper recommender system and an ontology containing information automatically extracted from departmental databases available on the web. The ontology is used to address the recommender systems cold-start problem. The recommender system addresses the ontology's interest-acquisition problem. An empirical evaluation of this approach is conducted and the performance of the integrated systems measured.

## 1. INTRODUCTION
The mass of content available on the World-Wide Web raises important questions over its effective use. Search engines filter web pages that match explicit queries, but most people find articulating exactly what they want difficult. The result is large lists of search results that contain a handful of useful pages, defeating the purpose of filtering in the first place.

Recommender systems [23] learn about user preferences over time and automatically find things of similar interest, thus reducing the burden of creating explicit queries. They dynamically track users as their interests change. However, such systems require an initial learning phase where behaviour information is built up to form an user profile. During this initial learning phase performance is often poor due to the lack of user information; this is known as the cold-start problem [17].

There has been increasing interest in developing and using tools for creating annotated content and making it available over the semantic web. Ontologies are one such tool, used to maintain and provide access to specific knowledge repositories. Such sources could complement the behavioral information held within recommender systems, by providing some initial knowledge about users and their domains of interest. It should thus be possible to bootstrap the initial learning phase of a recommender system with such knowledge, easing the cold-start problem.

In return for any bootstrap information the recommender system could provide details of dynamic user interests to the ontology. This would reduce the effort involved in acquiring and maintaining knowledge of people's research interests. To this end we investigate the integration of Quickstep, a web-based recommender system, an ontology for the academic domain and OntoCoPI, a community of practice identifier that can pick out similar users.

## 2. RECOMMENDER SYSTEMS
People may find articulating what they want hard, but they are good at recognizing it when they see it. This insight has led to the utilization of relevance feedback [24], where people rate web pages as interesting or not interesting and the system tries to find pages that match the interesting, positive examples and do not match the not interesting, negative examples. With sufficient positive and negative examples, modern machine learning techniques can classify new pages with impressive accuracy. Such systems are called content-based recommender systems.

Another way to recommend pages is based on the ratings of other people who have seen the page before. Collaborative recommender systems do this by asking people to rate explicitly pages and then recommending new pages that similar users have rated highly. The problem with collaborative filtering is that there is no direct reward for providing examples since they only help other people. This leads to initial difficulties in obtaining a sufficient number of ratings for the system to be useful.

Hybrid systems, attempting to combine the advantages of content-based and collaborative recommender systems, have proved popular to-date. The feedback required for content-based recommendation is shared, allowing collaborative recommendation as well. We use the Quickstep [18] hybrid recommender system in this paper to recommend on-line research papers.

## 2.1 The Cold-start Problem

One difficult problem commonly faced by recommender systems is the cold-start problem [17], where recommendations are required for new items or users for whom little or no information has yet been acquired. Poor performance resulting from a cold-start can deter user uptake of a recommender system. This effect is thus self-destructive, since the recommender never achieves good performance since users never use it for long enough. We will examine two types of cold-start problem.

The *new-system cold-start* problem is where there are no initial ratings by users, and hence no profiles of users. In this situation most recommender systems have no basis on which to recommend, and hence perform very poorly.

The *new-user cold-start* problem is where the system has been running for a while and a set of user profiles and ratings exist, but no information is available about a new user. Most recommender systems perform poorly in this situation too.

Collaborative recommender systems fail to help in cold-start situations, as they cannot discover similar user behaviour because there is not enough previously logged behaviour data upon which to base any correlations. Content-based and hybrid recommender systems perform a little better since they need just a few examples of user interest in order to find similar items.

No recommender system can cope alone with a totally cold-start however, since even content-based recommenders require a small number of examples on which to base recommendations. We propose to link together a recommender system and an ontology to address this problem. The ontology can provide a variety of information on users and their publications. Publications provide important information about what interests a user has had in the past, so provide a basis upon which to create initial profiles that can address the new-system cold start problem. Personnel records allow similar users to be identified. This will address the new-user cold-start problem by providing a set of similar users on which to base a new-user profile.

## 3. ONTOLOGIES

An ontology is a conceptualisation of a domain into a human-understandable, but machine-readable format consisting of entities, attributes, relationships, and axioms [12]. Ontologies can provide a rich conceptualisation of the working domain of an organisation, representing the main concepts and relationships of the work activities. These relationships could represent isolated information such as an employee's home phone number, or they could represent an activity such as authoring a document, or attending a conference.

In this paper we use the term ontology to refer to the classification structure and instances within the knowledge base.

The ontology used in our work is designed to represent the academic domain, and was developed by Southampton's AKT team (Advanced Knowledge Technologies [20]). It models people, projects, papers, events and research interests. The ontology itself is implemented in Protégé 2000 [10], a graphical tool for developing knowledge-based systems. It is populated with information extracted automatically from a departmental personnel database and publication database. The ontology consists of around 80 classes, 40 slots, over 13000 instances and is focused on people, projects, and publications.

## 3.1 The Interest-acquisition Problem

People's areas of expertise and interests are an important type of knowledge for many applications, for example expert finders [9]. Semantic web technology can be a good source of such information, but usually requires substantial maintenance to keep the web pages up-to-date. The majority of web pages receive little maintenance, holding information that does not date quickly. Since interests and areas of expertise are dynamic in nature they are not often held within web pages. It is thus particularly difficult for an ontology to acquire such information; this is the *interest-acquisition* problem.

Many existing systems force users to perform self-assessment to gather such information, but this has numerous disadvantages [5]. Lotus have developed a system that monitors user interaction with a document to capture interests and expertise [16]. Their system does not, however, consider the online documents that users browse.

This paper investigates linking an ontology with a recommender system to help overcoming the interest acquisition problem. The recommender system will regularly provide the ontology with interest profiles for users, obtained by monitoring user web browsing and analysing feedback on recommended research papers.

## 4. Related Work

Collaborative recommender systems utilize user ratings to recommend items liked by similar people. PHOAKS [26] is an example of a collaborative filtering, recommending web links mentioned in newsgroups articles. Only newsgroups with at least 20 posted web links are considered by PHOAKS, avoiding the cold-start problems associated with newer newsgroups containing less messages. Group Lens [14] is an alternative example, recommending newsgroup articles. Group Lens reports two cold-start problems in their experimental analysis. Users abandoned the system before they had provided enough ratings to receive recommendations and early adopters of the system received poor recommendations until enough ratings were gathered. These systems are typical of collaborative recommenders, where a cold-start makes early recommendation poor until sufficient people have provided ratings.

Content-based recommender systems recommend items with similar content to things the user has liked before. An example of a content-based recommender is Fab [4], which recommends web pages. Fab needs a few early ratings from each user in order to create a training set. ELFI [25] is another content-based recommender, recommending funding information from a database. ELFI observes users using a database and infers both positive and negative examples of interest from this behaviour. Both these systems are typical of content-based recommender systems, requiring users to use the system for an initial period of time before the cold-start problem is overcome.

Personal web-based agents such as Letizia [15], Syskill & Webert [21] and Personal Webwatcher [19] track the users browsing and formulate user profiles. Profiles are constructed from positive and negative examples of interest, obtained from explicit feedback or heuristics analysing browsing behaviour. They then suggest which links are worth following from the current web page by recommending page links most similar to the users profile. Just like a content-based recommender system, a few examples of interest must be observed or elicited from the user before a useful profile can be constructed.

Ontologies can be used to improve content-based search, as seen in OntoSeek [13]. Users of OntoSeek navigate the ontology in order to formulate queries. Ontologies can also be used to automatically construct knowledge bases from web pages, such as in Web-KB [8]. Web-KB takes manually labelled examples of domain concepts and applies machine-learning techniques to classify new web pages. Both systems do not, however, capture dynamic information such as user interests.

Also of relevance are systems such as CiteSeer [6], which use content-based similarity matching to help search for interesting research papers within a digital library.

# 5. THE QUICKSTEP RECOMMENDER SYSTEM

Quickstep [18] is a hybrid recommender system, addressing the real-world problem of recommending on-line research papers to researchers. User browsing behaviour is unobtrusively monitored via a proxy server, logging each URL browsed during normal work activity. A nearest-neighbour algorithm classifies browsed URL's based on a training set of labelled example papers, storing each new paper in a central database. The database of known papers grows over time, building a shared pool of knowledge. Explicit feedback and browsed URL's form the basis of the interest profile for each user. Figure 1 shows an overview of the Quickstep system.

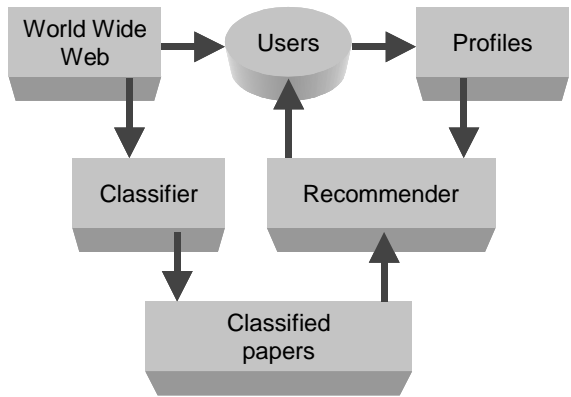

**Figure 1. The Quickstep recommender system**

Each day a set of recommendations is computed, based on correlations between user interest profiles and classified paper topics. Any feedback offered by users on these recommendations is recorded when the user looks at them. Users can provide new examples of topics and correct paper classifications where wrong. In this way the training set, and hence classification accuracy, improves over time.

Quickstep bases its user interest profiles on an ontology of research paper topics. This allows inferences from the ontology to assist profile generation; in our case topic inheritance is used to infer interest in super-classes of specific topics. Sharing interest profiles with the AKT ontology is not difficult since they are explicitly represented using ontological terms.

Previous trials [18] of Quickstep used hand-crafted initial profiles, based on interview data, to cope with the cold-start problem. Linking Quickstep with the AKT ontology automates this process, allowing a more realistic cold-start solution that will scale to larger numbers of users.

## 5.1 Paper classification algorithm

Every research paper within Quickstep's central database is represented using a term frequency vector. Terms are single words within the document, so term frequency vectors are computed by counting the number of times words appear within the paper. Each dimension within a vector represents a term. Dimensionality reduction on vectors is achieved by removing common words found in a stop-list and stemming words using the Porter [22] stemming algorithm. Quickstep uses vectors with 10-15,000 dimensions.

Once added to the database, papers are classified using an IBk [1] classifier boosted by the AdaBoostM1 [11] algorithm. The IBk classifier is a k-Nearest Neighbour type classifier that uses example documents, called a training set, added to a vector space. Figure 2 shows the basic k-Nearest Neighbour algorithm. The closeness of an unclassified vector to its neighbours within the vector space determines its classification.

$$w(d_a, d_b) = \sqrt{\sum_{j=1..T} (t_{ja} - t_{jb})^2}$$

$w(d_a, d_b)$    kNN distance between document a and b
$d_a, d_b$    document vectors
$T$    number of terms in document set
$t_{ja}$    weight of term j document a

**Figure 2. k-Nearest Neighbour algorithm**

Classifiers like k-Nearest Neighbour allow more training examples to be added to their vector space without the need to re-build the entire classifier. They also degrade well, so even when incorrect the class returned is normally in the right "neighbourhood" and so at least partially relevant. This makes k-Nearest Neighbour a robust choice of algorithm for this task.

Boosting works by repeatedly running a weak learning algorithm on various distributions of the training set, and then combining the classifiers produced by the weak learner into a single composite classifier. The "weak" learning algorithm here is the IBk classifier. Figure 3 shows the AdaBoostM1 algorithm.

```
Initialise all values of D to 1/N
Do for t=1..T
        call weak-learn(D_t)
        calculate error e_t
        calculate β_t = e_t/(1-e_t)
        calculate D_{t+1}
```

$$\text{classifier} = \underset{c \in C}{\operatorname{argmax}} \sum_{\substack{t = \text{all iterations} \\ \text{with result class } c}} \log \frac{1}{\beta_t}$$

| | |
|---|---|
| $D_t$ | class weight distribution on iteration t |
| N | number of classes |
| T | number of iterations |
| weak-learn($D_t$) | weak learner with distribution $D_t$ |
| $e_t$ | weak_learn error on iteration t |
| $\beta_t$ | error adjustment value on iteration t |
| classifier | final boosted classifier |
| C | all classes |

**Figure 3. AdaBoostM1 boosting algorithm**

AdaBoostM1 has been shown to improve [11] the performance of weak learner algorithms, particularly for stronger learning algorithms like k-Nearest Neighbour. It is thus a sensible choice to boost our IBk classifier.

## 5.2 User profiling algorithm

The profiling algorithm performs correlation between paper topic classifications and user browsing logs. Whenever a research paper is browsed that has been classified as belonging to a topic, it accumulates an interest score for that topic. Explicit feedback on recommendations also accumulates interest value for topics. The current interest of a topic is computed using the inverse time weighting algorithm shown in Figure 4.

$$\text{Topic interest} = \sum_{1..\text{no of instances}}^{n} \text{Interest value}(n) / \text{days old}(n)$$

Interest values  
Paper browsed = 1  
Recommendation followed = 2  
Topic rated interesting = 10  
Topic rated not interesting = -10

**Figure 4. Profiling algorithm**

An is-a hierarchy of research paper topics is held so that super-class relationships can be used to infer broader topic interest. When a specific topic is browsed, fractional interest is inferred for each super-class of that topic, using a $1/2^{\text{level}}$ weighting where 'level' refers to how many classes up the is-a tree the super-class is from the original topic. Figure 5 shows a section from the research paper topic ontology.

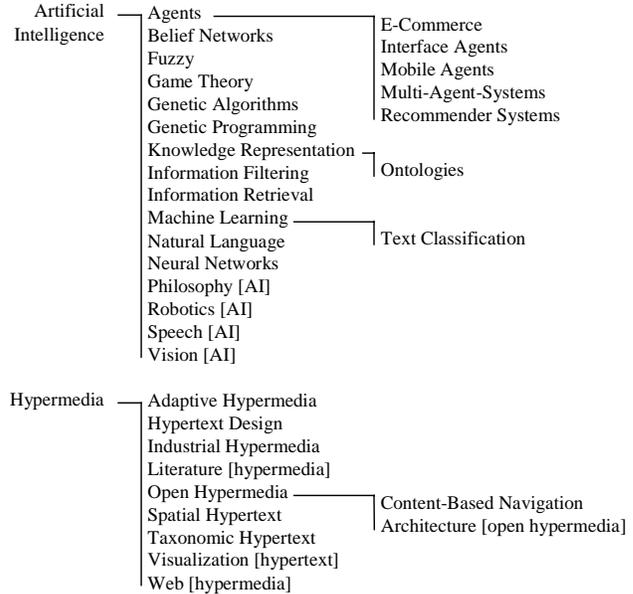

**Figure 5. Section of the research paper topic ontology**

## 5.3 Recommendation algorithm

Recommendations are formulated from a correlation between the users current topics of interest and papers classified as belonging to those topics. A paper is only recommended if it does not appear in the users browsed URL log, ensuring that recommendations have not been seen before. For each user, the top three interesting topics are selected with 10 recommendations made in total. Papers are ranked in order of the recommendation confidence before being presented to the user. Figure 6 shows the recommendation algorithm.

Recommendation confidence =   classification confidence *
                              topic interest value

**Figure 6. Recommendation algorithm**

## 6. ONTOCOPI

The Ontology-based Communities of Practice Identifier (OntoCoPI) [2] is an experimental system that uses the AKT ontology to help identifying communities of practice (CoP). The community of practice of a person is taken here to be the closest group of people, based on specific features they have in common with that given person. A community of practice is thus an informal group of people who share some common interest in a particular practice [7] [27]. Workplace communities of practice improve organisational performance by maintaining implicit knowledge, helping the spread of new ideas and solutions, acting as a focus for innovation and driving organisational strategy.

Identifying communities of practice is an essential first step to understand the knowledge resources of an organization [28]. Organisations can bring the right people together to help the identified communities of practice to flourish and expand, for example by providing them with appropriate infrastructure and give them support and recognition. However, community of practice identification is currently a resource-heavy process largely based on interviews, mainly because of the informal nature of such community structures that are normally hidden within and across organisations.

OntoCoPI is a tool that uses ontology-based network analysis to support the task of community of practice identification. A breadth-first spreading activation algorithm is applied by OntoCoPI to crawl the ontology network of instances and relationships to extract patterns of certain relations between entities relating to a community of practice. The crawl can be limited to a given set of ontology relationships. These relationships can be traced to find specific information, such as who attended the same events, who co-authored papers and who are members of the same project or organisation. Communities of practice are based on informal sets of relationships while ontologies are normally made up of formal relationships. The hypothesis underlying OntoCoPI is that some informal relationships can be inferred from the presence of formal ones. For instance, if A and B have no formal relationships, but they have both authored papers with C, then that could indicate a shared interest.

One of the advantages of using an ontology to identify communities of practice, rather than other traditional information networks [3] is that relationships can be selected according to their semantics, and can have different weights to reflect relative importance. For example the relations of document authorship and project membership can be selected if it is required to identify communities of practice based on publications and project work. OntoCoPI allows manual selection of relationships or automatic selection based on the frequency of relationship use within the knowledge base. Selecting the right relationships and weights is an experimental process that is dependent on the ontology structure, the type and amount of information in the ontology, and the type of community of practice required.

When working with a new community of practice some experiments will be needed to see which relationships are relevant to the desired community of practice, and how to set relative weights. In the experiments described in this paper, certain relationships were selected manually and weighted based on our preferences. Further trials are needed to determine the most effective selection.

## 7. INTEGRATION OF THE TWO TECHNOLOGIES

We have investigated the integration of the ontology, OntoCoPI and Quickstep recommender system to provide a solution to both the cold-start problem and interest acquisition problem. Figure 7 shows our experimental systems after integration.

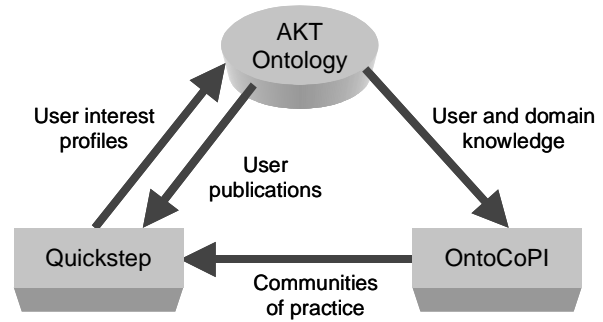

**Figure 7. Ontology and recommender system integration**

Upon start-up, the ontology provides the recommender system with an initial set of publications for each of its registered users. Each user's known publications are then correlated with the recommender systems classified paper database, and a set of historical interests compiled for that user. These historical interests form the basis of an initial profile, overcoming the new-system cold-start problem. Figure 8 details the initial profile algorithm. As per the Quickstep profiling algorithm, fractional interest in a topic super-classes is inferred when a specific topic is added.

$$\text{topic interest}(t) = \sum_{\substack{1..\text{ publications} \\ \text{belonging to class } t}}^{n} 1 / \text{publication age}(n)$$

new-system initial profile = (t, topic interest(t))*

t = <research paper topic>

**Figure 8. New-system initial profile algorithm**

When the recommender system is up and running and a new user is added, the ontology provides the historical publication list of the new user and the OntoCoPI system provides a ranked list of similar users. The initial profile of the new user is formed from a correlation between historical publications and any similar user profiles. This algorithm is detailed in figure 9, and addresses the new-user cold-start problem.

$$\text{topic interest}(t) = \frac{\gamma}{N_{similar}} \sum_{1..N_{similar}}^{u} \text{profile interest}(u,t)$$

$$+ \sum_{1..N_{pubs\ t}}^{n} 1 / \text{publication age}(n)$$

profile interest(u,t) = interest of user u in topic t * CoP confidence
new-user initial profile = (t, topic interest(t))*

    t = research paper topic
    u = user
    γ = weighting constant >= 0
    $N_{similar}$ = number of similar users
    $N_{pubs\ t}$ = number of publications belonging to class t
    CoP confidence = Communities of practice confidence

**Figure 9. New-user initial profile algorithm**

The task of populating and maintaining the ontology of user research interests is left to the recommender system. The recommender system compiles user profiles on a daily basis, and these profiles are asserted into the ontology when ready. Figure 10 details the structure of these profiles. In this way up-to-date interests are maintained, providing a solution to the interest acquisition problem. The interest data is used alongside the more static information within the ontology to improve the accuracy of the OntoCoPI system.

user profile = (topic, interest)*
topic = research topic
interest = interest value

**Figure 10. Daily profiles sent to the AKT ontology**

## 7.1 Example of system operation

When the Quickstep recommender system is first initialised, it retrieves a list of people and their publication URLs from the ontology. Quickstep analyses these publications and classifies them according to the research topic hierarchy in the ontology. Paper topics are associated with their date of publication, and the 'new-system initial profile' algorithm used to compute a set of initial profiles for each user.

Tables 1 and 2 shows an example of this for the user Nigel Shadbolt. His publications are analysed and a set of topics and dates formulated. The 'new-system initial profile' algorithm then computes the interest values for each topic. For example, 'Knowledge Acquisition' has one publication two year old (round up) so its value is 1.0 / 2 = 0.5.

**Table1. Publication list for Shadbolt**

| Publication | Date | Topic |
|---|---|---|
| Capturing Knowledge of User Preferences: ontologies on recommender systems | 2001 | Recommender systems |
| Knowledge Technologies | 2001 | Knowledge Technology |
| The Use of Ontologies for Knowledge Acquisition | 2001 | Ontology |
| Certifying KBSs: Using CommonKADS to Provide Supporting Evidence for Fitness for Purpose of KBSs | 2000 | Knowledge Management |
| Extracting Focused Knowledge from the Semantic Web | 2000 | Knowledge Acquisition |
| Knowledge Engineering and Management | 2000 | Knowledge Management |
| … | | |

**Table2. Example of new-system profile for Shadbolt**

| Topic | Interest |
|---|---|
| Knowledge Technology\Knowledge Management | 1.5 |
| Knowledge Technology\Ontology | 1.0 |
| AI\Agents\Recommender Systems | 1.0 |
| Knowledge Technology\Knowledge Acquisition | 0.5 |
| … | |

At a later stage, after Quickstep has been running for a while, a new user registers with email address sem99r@ecs.soton.ac.uk. OntoCoPI identifies this email account as that of Stuart Middleton, a PhD candidate within the department, and returns the ranked and normalised communities of practise list displayed in table 3. This communities of practise list is identified using relations on conference attendance, supervision, authorship, research interest, and project membership, using the weights 0.4, 0.7, 0.3, 0.8, and 0.5 respectively. De Roure was found to be the closest person as he is Middleton's supervisor, and has one joint publication co-authored with Middleton and Shadbolt. The people with 0.82 values are other supervisees of De Roure. Alani attended the same conference that Middleton went to in 2001.

**Table 3. OntoCoPI results for Middleton**

| Person | Relevance value | Person | Relevance value |
|---|---|---|---|
| DeRoure | 1.0 | Alani | 0.47 |
| Revill | 0.82 | Shadbolt | 0.46 |
| Beales | 0.82 | | |

The communities of practise list is then sent to Quickstep, which searches for matching user profiles. These profiles will be more accurate and up to date than those initially created profiles based on publications. Quickstep manages to find the profiles in table 4 in its logs.

**Table 4. Profiles of similar people to Middleton**

| Person | Topic | Interest |
|---|---|---|
| DeRoure | AI\Distributed Systems | 1.2 |
| | AI\Agents\Recommender Systems … | 0.73 |
| Revill | AI\Agents\Mobile Agents | 1.0 |
| | AI\Agents\Recommender Systems … | 0.4 |
| Beals | Knowledge Technology\Knowledge Devices | 0.9 |
| | AI\Agents\Mobile Agents … | 0.87 |
| Alani | Knowledge Technology\Ontology | 1.8 |
| | Knowledge Technology\Knowledge Management\ CoP … | 0.7 |
| Shadbolt | Knowledge Technology\Knowledge Management | 1.5 |
| | AI\Agents\Recommender Systems … | 1.0 |

These profiles are merged to create a profile for the new user, Middleton, using the 'new-user initial profile' algorithm with a γ value of 2.5. For example, Middleton has a publication on 'Recommender Systems' that is 1 year old and DeRoure, Revill and Shadbolt have interest in 'Recommender Systems' – this topics value is therefore 1/1 + 2.5/5 * (1.0*0.73+0.82*0.4+0.46*1.0) = 1.76. Table 5 shows the resulting profile.

**Table 5. New-user profile for Middleton**

| Topic | Interest |
|---|---|
| AI\Agents\Recommender Systems | 1.76 |
| AI\Agents\Mobile Agents | 0.77 |
| AI\Distributed Systems | 0.6 |
| Knowledge Technology\Ontology | 0.42 |
| Knowledge Technology\Knowledge Devices | 0.37 |
| Knowledge Technology\Knowledge Management | 0.35 |
| Knowledge Technology\Knowledge Management\ CoP | 0.16 |
| … | |

Every day Quickstep's profiles are updated and automatically fed back to the ontology, where they are used to populate the research interest relationships of the relevant people.

## 8. EMPIRICAL EVALUATION

In order to evaluate the effect both the new-system and new-user initial profiling algorithms have on our integrated system, we conducted an experiment based around the browsing behaviour logs obtained from the Quickstep [18] user trials. The algorithms previously described are used, as per the example in the previous section, and the average performance for all users calculated.

### 8.1 Experimental approach

Users were selected from the Quickstep trials whom had entries within the departmental publication database. We selected nine users in total, with each user typically having one or two publications.

The URL browsing logs of these users, extracted from 3 months of browsing behaviour recorded during the Quickstep trials, were then broken up into weekly log entries. Seven weeks of browsing behaviour were taken from the start of the Quickstep trials, and an empty log created to simulate the very start of the trial.

Eight iterations of the integrated system were thus run, the first simulating the start of the trial and others simulating the following weeks 1 to 7. Interest profiles were recorded after each iteration. Two complete runs were made, one with the 'new-system initial profiling' algorithm and one without. The control run without the 'new-system initial profiling' algorithm started with blank profiles for each of its users.

An additional iteration was run to evaluate the effectiveness of the 'new-user initial profile' algorithm. We took the communities of practice for each user, based on data from week 7, and used the 'new-user initial profile' algorithm to compute initial profiles for each user as if they were being entered onto the system at the end of the trial. These profiles were recorded. Because we are using an early prototype version of OntoCoPI, communities of practice confidence values were not available; we thus used confidence values of 1 throughout this experiment.

In order to evaluate our algorithms effect on the cold-start problem, we compared all recorded profiles to the benchmark week 7 profile. This allows us to measure how quickly profiles converge to the stable state existing after a reasonable amount of behaviour data has been accumulated. The quicker the profiles move to this state the quicker they will have overcome the cold-start.

Week 7 was chosen as the cut-off point of our analysis since after about 7 weeks of use the behaviour data gathered by Quickstep will dominate the user profiles. The effects of bootstrapping beyond this point would not be significant. If we were to run the system beyond week 7 we would simply see the profiles continually adjusting to the behaviour logged each week.

## 8.2 Experimental results

Two measurements were preformed when comparing profiles to the benchmark week 7 profile. The first, profile precision, measures how many topics were mentioned in both the current profile and benchmark profile. Profile precision is an indication of how quickly the profile is converging to the final state, and thus how quickly the effects of the cold-start are overcome. The second, profile error rate, measures how many topics appeared in the current profile that did not appear within the benchmark profile. Profile error rate is an indication of the errors introduced by the two bootstrapping algorithms. Figure 11 describes these metrics.

It should be noted that we are not measuring the absolute precision and error rate of the profiles – only the relative precision and error rate compared to the week 7 steady state profiles. Measuring absolute profile accuracy is a very subjective matter, and we do not attempt it here; we are only interested in how quickly profiles reach their steady states. A more complete evaluation of Quickstep's overall profiling and recommendation performance can be found in [18].

$$\text{profile precision} = \frac{1}{N_{users}} \sum_{1..N_{users}}^{user} \frac{N_{correct}}{N_{correct} + N_{missing}}$$

$$\text{profile error rate} = \frac{1}{N_{users}} \sum_{1..N_{users}}^{user} \frac{N_{incorrect}}{N_{correct} + N_{incorrect} + N_{missing}}$$

$N_{correct}$  Number of user topics that appear in current profile and benchmark profile
$N_{missing}$  Number of user topics that appear in benchmark profile but not in current profile
$N_{incorrect}$ Number of user topics that appear in current profile but not in benchmark profile
$N_{users}$   Total number of users

**Figure 11. Evaluation metrics**

The results of our experimental runs are detailed in figures 12 and 13. The new-user results consist of a single iteration, so appear on the graphs as a single point.

At the start, week 0, no browsing behaviour log data is available to the system so the profiles without bootstrapping are empty. The new-system algorithm, however, can bootstrap the initial user profiles and achieves a reasonable precision of 0.35 and a low error rate of 0.06. We found that the new-system profiles accurately captured interests users had a year or so ago, but tended to miss current interests. This is because publications are generally not available for up-to-date interests.

As we would expect, once the weekly behaviour logs become available to the system the profiles adjust accordingly, moving away from the initial bootstrapping. On week 7 the profiles converge to the benchmark profile.

The new-user algorithm result show a more dramatic increase in precision to 0.84, but comes at the price of a significant error rate of 0.55. The profiles produced by the new-user algorithm tended to be very inclusive, taking the set of similar user interests and producing a union of these interests. While this captures many of the new users real interests, it also included a large number of interests not relevant to the new user but which were interesting to the people similar to the new user.

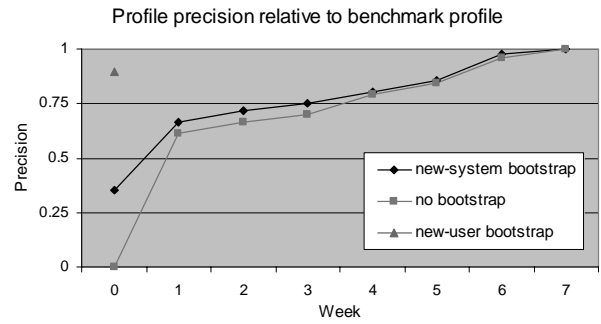

**Figure 12. Profile precision**

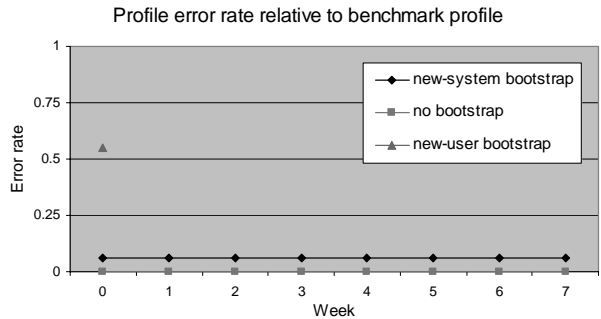

**Figure 13. Profile error rate**

Since error rate is measured relative to the final benchmark profile of week 7, all the topics seen in the behaviour logs will be present within the benchmark profile. Incorrect topics must thus come from another source – in this case bootstrapping on week 0. This causes error rates to be constant over the 7 weeks, since the incorrect topics introduced on week 0 remain for all subsequent weeks.

## 9. DISCUSSION

Cold-starts in recommender systems and interest acquisition in ontologies are serious problems. If initial recommendations are inaccurate, user confidence in the recommender system may drop with the result that not enough usage data is gathered to overcome the cold-start. In regards to ontologies, up-to-date interests are not generally available from periodically updated information sources such as web pages, personal records or publication databases.

Our integration of the Quickstep recommender system, AKT ontology and OntoCoPI system has demonstrated one approach to reduce both the cold-start and interest-acquisition problems. Our practical work suggests that using an ontology to bootstrap user profiles can significantly reduce the impact of the recommender system cold-start problem. It is particularly useful for the new-system cold-start problem, where the alternative is to start with no information at all. Regularly feeding the recommender systems interest profiles back to the ontology also clearly assists in the acquisition of up-to-date interests. A number of issues have, however, arisen from our integration.

The new-system algorithm produced profiles with a low error rate and a reasonable precision of 0.35. This reflects that previous publications are a good indication of users current interests, and so can produce a good starting point for a bootstrap profile. Where the new-system algorithm fails is for more recent interests, which make up the remaining 65% of the topics in the final benchmark profile. To discover these more recent interests, it is possible that the new-system algorithm could be extended to take some of the other information available within the ontology into account, such as the projects a user is working on. To what degree these relationships will help is difficult to predict however, since the ontology itself has great difficulty in acquiring knowledge of recent interests.

For the purposes of our experiment, we selected those users who had some entries within the universities on-line publication database. There were some users who had not entered their publications into this database or who have yet to publish their work. For these users there is little information within the ontology, making any new-system initial profiles of little use. In a larger scale system, more sources of information would be needed from the ontology to build the new-system profiles. This would allow some redundancy, and hence improve robustness in the realistic situation where information is sparsely available.

The community of practice for a user was found not to be always relevant based on our selection of relationships and weights. For example, Dave de Roure supervises Stuart Middleton, but Dave supervises a lot of other students interested in mobile agents. These topics are not relevant to Stuart, which raises the question of how relevant the supervision relationship is to our requirements, and how best to weight such a relationship. Further experiments are needed to identify the most relevant settings for community of practice identification. The accuracy of our communities of practice are also linked to the accuracy of the research interest information as identified by the recommender system.

The new-user algorithm achieved good precision of 0.84 at the expense of a significant 0.55 error rate. This was because both the communities of practice generated for users were not always precise, and because of the new-user algorithm included all interests from the similar users. An improvement would be to only use those interests held by the majority of the people within a community of practice. This would exclude some of the less common interests that would otherwise be included into the new-user profile.

The new-user initial profile algorithm defines the constant $\gamma$, which determines the proportional significance of previous publications and similar users. Factors such as the availability of relationship data within the ontology and quality of the publication database will affect the choice of value for $\gamma$. We used a value of 2.5, but empirical evaluation would be needed to determine the best value.

There is an issue as to how best to calculate the "semantic distance" between topics within the is-a hierarchy. We make the simplifying assumption that all is-a links have equal relevance, but the exact relevance will depend on each topic in question. If individual weightings were allowed for each topic, a method for determination of these weights would have to be considered. Alternatively the is-a hierarchy could be carefully constructed to ensure equal semantic distance.

A positive feedback loop exists between the recommender system and ontology, making data incest a potential problem. For new users there are no initial interest entries within the ontology, so new user profiles are not incestuous. If the recommender system were to use the communities of practice for more than just initial profiles, however, a self-confirming loop would exist and interest calculations would be incestuous.

Finally, a question still remains as to just how good an initial profile must be to fully overcome the effects of the cold-start problem. If initial recommendations are poor users will not use the recommender system and hence it will never get a chance to improve. We have shown that improvements can be made to initial profiles, but further empirical evaluation would be needed to evaluate exactly how much improvement is needed before the system is "good enough" for users to give it a chance.

## 10. FUTURE WORK

The next step for the integrated system is to continue to improve the set of relationships and weights used to calculate communities of practice, and find a more selective 'new-user initial profile' algorithm. With more precise communities of practice the new-user bootstrapping error rate should fall substantially. We could then conduct a set of further user trials. This would allow the assessment of user up-take and use of the integrated system, and reveal how improving initial profiles affect overall system usage patterns.

The Quickstep recommender system is currently being extended to explore further the idea of using ontologies to represent user profiles. A large-scale trial is under way over a full academic year to evaluate the new system, which is called the Foxtrot recommender system.

## 11. ACKNOWLEDGEMENTS
This work is funded by EPSRC studentship award number 99308831 and the AKT project.